\documentclass[sigconf,anonymous=false]{acmart}
\AtBeginDocument{%
  \providecommand\BibTeX{{%
    \normalfont B\kern-0.5em{\scshape i\kern-0.25em b}\kern-0.8em\TeX}}}

\setcopyright{acmcopyright}

\begin{document}
\copyrightyear{2018}
\acmYear{2018}
\acmDOI{10.1145/1122445.1122456}
\acmConference[Woodstock '18]{Woodstock '18: ACM Symposium on Neural
  Gaze Detection}{June 03--05, 2018}{Woodstock, NY}
\acmBooktitle{Woodstock '18: ACM Symposium on Neural Gaze Detection,
  June 03--05, 2018, Woodstock, NY}
\acmPrice{15.00}
\acmISBN{978-1-4503-XXXX-X/18/06}

\title{AdaCoach: A Virtual Coach for Training Customer Service Agents}

\author{Shuang Peng, Shuai Zhu, Minghui Yang, Haozhou Huang\\ Dan Liu, Zujie Wen, Xuelian Li, Biao Fan}
\affiliation{
  \institution{Ant Group}
  \city{HangZhou}
  \country{China}}
\email{{jianfeng.ps, zs261988, minghui.ymh, haozhou.hhz, yuewei.ld, zujie.wzj, yiru.lxl, biao.fan}@antgroup.com}








\renewcommand{\shortauthors}{Peng, et al.}

\begin{abstract}
With the development of online business, customer service agents gradually play a crucial role as an interface between the companies and their customers. Most companies spend a lot of time and effort on hiring and training customer service agents. To this end, we propose \textbf{AdaCoach: A Virtual Coach for Training Customer Service Agents}, to promote the ability of newly hired service agents before they get to work. AdaCoach is designed to simulate real customers who seek help and actively initiate the dialogue with the customer service agents. 
Besides, AdaCoach uses an automated dialogue evaluation model to score the performance of the customer agent in the training process, which can provide necessary assistance when the newly hired customer service agent encounters problems.
We apply recent NLP technologies to ensure efficient run-time performance in the deployed system. 
To the best of our knowledge, this is the first system that trains the customer service agent through human-computer interaction. Until now, the system has already supported more than 500,000 simulation training and cultivated over 1000 qualified customer service agents.
\end{abstract}



\maketitle

\section{Introduction}
  With the development of online business, customer service has been applied in various domains, including technical support, after-sales service, and banking applications.
  Customer service is one of the pillars for a company's success, as it is highly related to customers' satisfaction and affects how the company is viewed by the public \cite{fadnis2020agent}. 
  To some extent, providing satisfying customer service can generate more marketing and sales opportunities. 
  For most business scenes, a newly hired customer service agent must receive a series of well-designed training courses and pass the certification exams before serving real customers. 
  
  The majority of the training process contains a part called practice rounds. In this part, the experienced customer service agent acts as a coach to simulate real customer who has problems and actively seeks help from the new customer agent. The new customer agent needs to determine and solve the problem precisely.  
  Based on the new customer service agent's performance during the practice rounds, the company administrator can be aware of the current vocational level of new customer agents and decide whether they are qualified for serving real online customers. 

  However, the traditional human-human training mode is always time-consuming and requires many human resources. In the customer service scenario, the shortage of experienced coaches leads to the fact that many newly hired customer service agents need to wait a long time for the opportunities of practice rounds and thus need a longer time to be qualified for serving the real customers.
  To this end, we propose an intelligent system called AdaCoach that aims at training newly hired customer service agents through the form of human-computer interaction and helping them quickly and better master the necessary service skills.
  For these purposes, we use recent NLP technologies, such as streaming automatic speech recognition (ASR)~\cite{moritz2020streaming, mani2020asr}, large scale pre-trained language models~\cite{devlin2018bert, chinesebertwwm}, text matching methods~\cite{chen2017enhanced, enhancedrcnn}, dialogue generation\cite{zhang2019dialogpt}, and dialogue evaluation~\cite{li2021dialoflow}, to ensure high performance. 

  \begin{figure*}[ht]
    \centering
    \includegraphics[width=1.6\columnwidth]{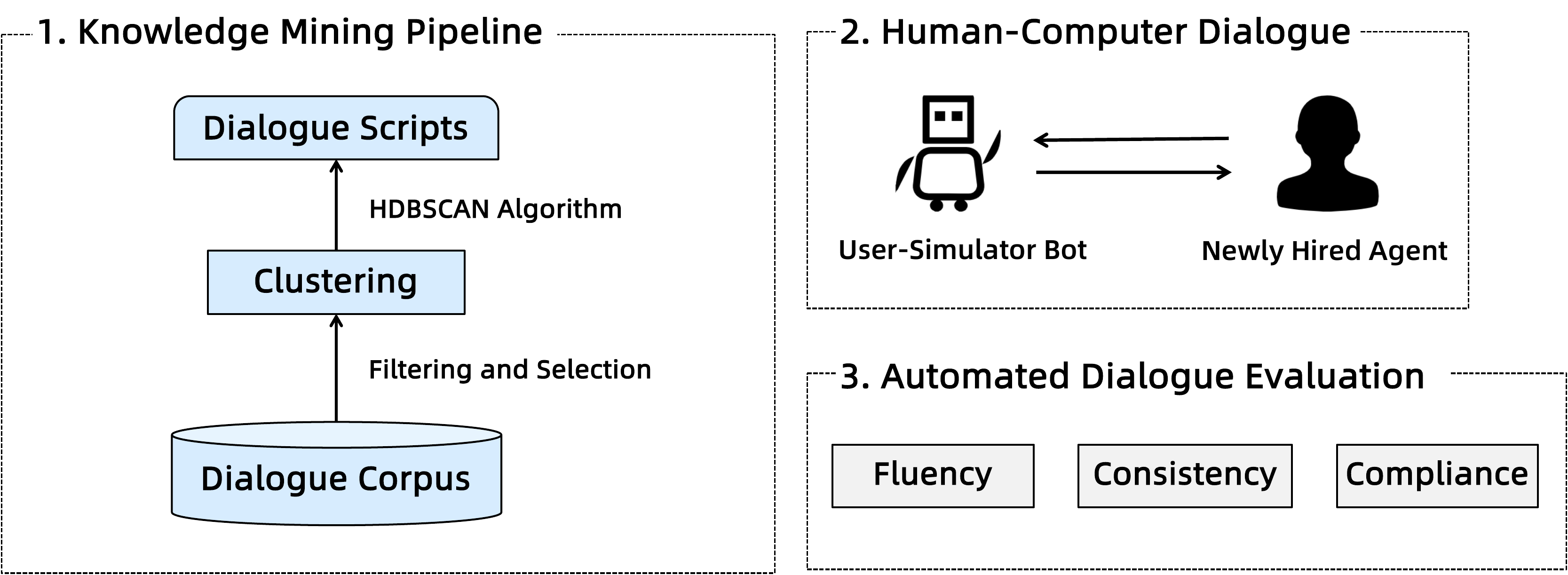}
    \caption{The System Architecture of AdaCoach.}
    \label{flow}
  \end{figure*}

  To the best of our knowledge, this is the first work to propose an intelligent system for training new customer service agents. The main contributions of our work are concluded as follows.
  \begin{itemize}
  \item We develop a knowledge mining pipeline for collecting enough dialogue corpus for training the downstream dialogue generation and dialogue evaluation models. 
  \item We propose a user-simulator dialogue bot that can simulate different types of customers in varied scenes.
  The dialogue bot can provide necessary assistance during the human-computer dialogue when the customer service agent stucks on some problems.
  \item We design an automated dialogue evaluation model for scoring the performance of customer service agents in the training process. The score can be used to decide whether the newly hired customer service agent is qualified for serving real customers.
  \end{itemize}

  \section{System Description}
  We present AdaCoach, an intelligent system that combines some dialogue-related technologies for helping the new customer service agent master the necessary skills and speeding up the whole training process.

  The illustration of the system is shown in Figure~\ref{flow}.
  The whole system is built by integrating several independent components such as a knowledge mining pipeline, a human-computer dialogue system, and an automated dialogue evaluation. 
  We introduce the detail of each component in the following subsections.
  
  \subsection{Knowledge Mining Pipeline}
  The knowledge mining pipeline is to select high-quality representative dialogue scripts\footnote{The dialogue script represents the collection of online service records (multi-turn dialogues) that customer service agent solves the customer's questions in varied scenes. These dialogue scripts are used to construct the dialogue system introduced in subsection 2.2.} from large-scale human-to-human conversation logs about customer service.
  These conversation logs are used as the training data of the human-computer dialogue model.
  In order to promote the quality of the knowledge mining pipeline and select dialogue scripts, we used the HDBSCAN algorithm \cite{mcinnes2017hdbscan} for clustering the original dialogue corpus into different user intents (e.g. complaint handling or product-use). These user intent clusters contain representative dialogue scripts in varied scenes. We utilize these dialogue scripts for building downstream user-simulator dialogue bots.

  \begin{figure*}[ht]
    \centering
    \includegraphics[width=1.6\columnwidth]{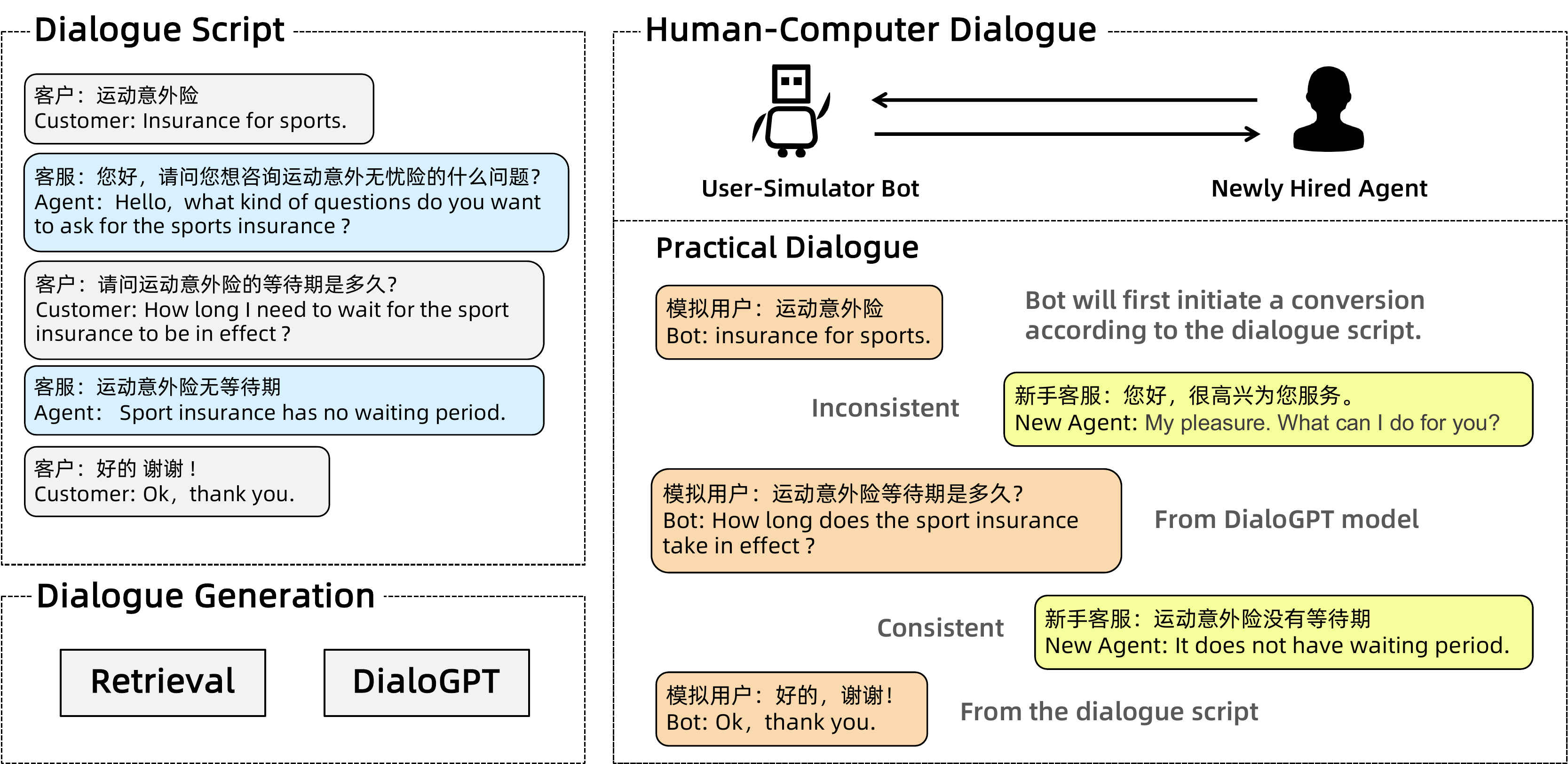}
    \caption{An illustration of the workflow of the Human-Computer Dialogue System. In this example, the user-simulator bot first initiates a conversation with the customer agent according to the dialogue script on a certain scene. Then the customer agent begins to solve the user's problems. The user-simulator bot can generate different user utterances from the existing dialogue scripts or dialogue generation models based on the customer agent's responses.}
    \label{exp}
  \end{figure*}

  \subsection{Human-Computer Dialogue System}
  The human-computer dialogue system is the core component of AdaCoach. 
  Like the traditional human-human training mode, the newly hired customer service agent improves their vocational level by continuously practicing with the user-simulator dialogue bot. The dialogue bot is designed to simulate customers who have problems in varied scenes and seek help from the customer service agent.
  Limited by the complexity of the customer service scene, the standard dialogue decision tree such as Google Dialogflow~\cite{sabharwal2020introduction} can not be directly applied to our scene. We adopted an alternative approach as follows.

  (1) The user-simulator bot first initiates the conversation, proposing a problem according to the dialogue script in a certain scene.

  (2) The new customer service agent will solve the problem proposed by the user-simulator bot based on the standard service procedure.

  (3) If the new customer service agent's response is similar to the corresponding utterance in the dialogue script, the human-computer dialogue will proceed. And if the agent's response is not similar to that in the dialogue script, the user-simulator bot will generate a response by the dialogue model.

  We show an example to illustrate the workflow in Figure~\ref{exp}. In order to best enhance the dialogue bot's user experience, we take both the retrieval and generation technologies for generating candidate responses.
  We implement the retrieval-based dialogue model referring to \cite{boussaha2019deep}. First, we encode the dialogue context into 200-dimensions vectors by the word2vec model \cite{pautovefficient}, and construct the mapping relation of the dialogue context and the corresponding user response. Then, we use the open-source framework Faiss~\cite{JDH17} to build a vector index and do a similarity search.
  Each time the ASR result is passed, the dialogue context is transported to the similarity search module.
  After identifying the top-3 similar dialogue context, we get the corresponding candidate user responses.
  
  Meanwhile, we implement the generation-based dialogue model following DialoGPT~\cite{zhang2019dialogpt}. We train different dialogue models on the different corpus based on the obtained dialogue corpus tagged with different user intents. For example, we select the dialogue corpus on the \textsl{complaint handling} scene for simulating customers who make a complaint about the product, and the customer service agent can practice with the dialogue bot for mastering skills how to solve customer problems about the product complaint. The generation model will generate three candidate responses given the current dialogue context.
  
  The candidate responses from the previous two parts are passed to the response-ranking module that consists of a binary dialogue BERT model~\cite{ijcai2020-639,chinesebertwwm}, calculating the reasonableness score of each response given the dialogue context. From the experimental results, the BLEU-2 of the response generation model achieves 42\%.

  \subsection{Automated Dialogue Evalution}
  We design an automated dialogue evaluation model for scoring the performance of the customer service agent during the training process. The score is given in the following three dimensions.

  \begin{itemize}
  \item \textbf{Fluency:} We use the Flow Score proposed in DialoFlow~\cite{li2021dialoflow} to measure the fluency of each turn's agent response and the overall fluency is the average of each turn. The score is from 0 to 1.
  \item \textbf{Consistency:} We use the text matching method~\cite{enhancedrcnn} to calculate the similarity between the new customer agent's responses and standard utterances in the dialogue script. The similarity threshold is set to 0.5, and the proportion of similar utterances during the training process represents the consistency score.
  The score is from 0 to 1.
  \item \textbf{Compliance:} We use the rule-based inspection to score the compliance of the dialogue. The score is either 0 or 1 based on the inspection of whether any agent response violates the service principles. 
  \end{itemize}

  The final score consists of 35\% of \textsl{Fluency}, 35\% of \textsl{Consistency}, and 30\% of \textsl{Compliance}.
  The system will display the scoring results based on the real-time dialogue context. If the new customer agent gets a low score on the previously introduced three dimensions, the scoring results would list detailed reasons that can assist them.
  We compare the system score with the artificial score, and the Top 100's Pearson correlation coefficient achieves 0.83.

  \section{Experiments}
  \subsection{Datasets}
  \begin{table}[]
    \centering
    \begin{tabular}{l|rr}
    \toprule
    \textbf{Items}               & \textbf{Training}            & \textbf{Validation}  \\ \hline
    Num of dialogs               & 500000                       & 3000            \\ 
    Average Rounds Per Dialog    & 31.6                         & 29.2          \\ 
    Average Length Per Dialog    & 603                          & 579          \\ 
    \bottomrule
    \end{tabular}
    \caption{Statistics of Agent Training Dataset.}
    \label{table:adacoach}
  \end{table}
  We manually construct an agent training dataset from the real-world environment and use it for evaluating the performance of AdaCoach. 
  The training and validation dataset consists of 500000 and 3000 dialogs from the real customer service scenario. 
  The training and validation datasets are used to train the DialoGPT and Flow Score models. Here we refer the experiment setting in \cite{zhang2019dialogpt,li2021dialoflow}. Due to space limitations, we do not report the separate experimental results on these two models.
  Table~\ref{table:adacoach} shows the detailed statistics of the agent training dataset.

  To protect data privacy, the usage of the customer's private data must be authorized. We have obtained explicit permission from the customer to use their private data. 
  If the customer disapproves of the authorization, we will not use their data. 
  For the authorized user privacy data, we have a series of strict processes to ensure the data remain confidential.

  \begin{table*}
  \centering
  \begin{tabular}{l|c|c|c|c}
    \toprule
  \textbf{Training Mode}&\textbf{Waiting Time (secs)}&\textbf{Average Durations (mins)}&\textbf{Average Rounds}&\textbf{Completion Rate (\%)}\\
    \hline
    Human-Human&308.5&7.3&25.6&81.4\\
    Human-Computer&3.1&6.8&20.8&80.2\\
  \bottomrule
  \end{tabular}
  \caption{The comparision results between Human-Human and Human-Computer training modes.}
  \label{table:agent_training_exp1}
  \end{table*}

  \subsection{Method in Comparison}
  As far as we know, there is no similar work in industry or academia. 
  To examine the effectiveness of AdaCoach, we compare it with the traditional Human-Human mode on the performance of agent training. We compare 1000 training records of Human-Human and Human-Computer modes of the same time period.

  Different from AdaCoach which is mainly based on the user-simulator dialogue bot, the traditional Human-Human training mode is mainly conducted by the experienced customer service agent who simulates as customers. They train the new customer service agents by asking some well-designed questions. These questions are proposed during the Human-Human conversation, and the answer qualities are the basis of evaluating the qualifications of the new customer service agent.

  \subsection{Evaluation Metrics}
  The evaluation process is conducted online.
  Because our work's primary purpose is to promote the efficiency of agent training and reduce the whole time cost, 
  we choose the following three dimensions as the evaluation metrics.
  \begin{enumerate}
  \item \textbf{Waiting Time:} The average time cost that the customer service agent waits to be assigned to a certain human coach or user-simulator bot. 
  \item \textbf{Average Durations:} The average training durations of the customer training process. 
  \item \textbf{Average Rounds:} The average rounds per dialog in the customer training process. 
  \item \textbf{Completion Rate:} The ratio of the new customer agent that correctly completes the standard service procedure. This is an important metric that measures the training quality.
  \end{enumerate}

  \subsection{Evaluation Results and Discussions} 
  The experimental results on the agent training performance are shown in Table~\ref{table:agent_training_exp1}.
  We summarize our observations as follows: 

  (1) Our Human-Computer training mode achieves competitive performance on completion rate with the traditional Human-Human training mode. This shows that using the user-simulator bot as the training coach is more effective with less dependent on the experienced coach.

  (2) Compared with the traditional Human-Human training mode, the waiting time of the Human-Computer training mode is only one-hundredth of the Human-Human training mode. This shows that our Human-Computer training mode effectively solves the problem of getting enough experienced coaches in traditional Human-Human training mode.

  (3) The training duration and average rounds per dialog of the Human-Computer training mode are less than the Human-Human training mode. This is because complex customer problems are always time-consuming, and the Human-Human training mode will spend more time and effort on these parts of customer cases. As a result, the Human-Human training mode is better than the Human-Computer training mode in solving complex problems.

  \subsection{Deployment Details}
  Our AdaCoach has been deployed in the real customer service scenario for six months and has cultivated more than 1000 qualified customer service agents in total. AdaCoach can support a peak inflow of 60 QPS.

  Based on statistical results, the average time cost of the training process is reduced from 25 days to 20 days, and now 80\% of the human-human training mode is replaced with the human-computer mode.
  This indicates that AdaCoach has effectively helped the new customer service agent. 
  
  \section{Related Work}
  With the advances in NLP, human-computer interaction (HCI) systems have been used in various domains~\cite{fadnis2020agent}. In this section, we briefly introduce some related technologies used in AdaCoach.

  \subsection{Pre-trained Models for Dialogue Generation}
  Recent advances in pre-trained language models have had great success in dialogue generation. DialoGPT~\cite{zhang2019dialogpt}, Plato-2~\cite{bao2020plato}, and DialoFlow~\cite{li2021dialoflow} achieve strong generation performances by training transformer-based language models on open-domain conversation corpus. 
  We have experimented with the above models and finally took DialoGPT into the AdaCoach based on the performance of dialogue generation responses.

  \subsection{Interactive Dialogue Evaluation}
  Interactive dialogue evaluation is a challenging problem, as there is no gold reference for the utterances. \cite{mehri2020unsupervised} propose the FED score, an automatic dialogue evaluation metric using the pre-trained model. 
  However, the FED score has limited performance on those dialogues without apparent comments. ~\cite{li2021dialoflow} further propose the Flow score, which entirely depends on the pre-trained model with no human integration.
  We take the Flow score~\cite{li2021dialoflow} into the dialogue evaluation of AdaCoach, measuring the fluency score.

  \subsection{Text Matching}
  Text Matching is a fundamental and important NLP task. In particular, a good model should also have the capacity to learn sentence similarity regardless of the length of the text and also needs to be efficient when applied to real-world applications~\cite{chen2017enhanced, enhancedrcnn}.

  In the scene of agent training, text matching is used to calculate the consistency score. Therefore we use the recent proposed Enhanced-RCNN model~\cite{enhancedrcnn} that achieves good performance on Chinese paraphrase identification datasets. 

  \section{Conclusions and Future Work}
  In this work, we present AdaCoach, an intelligent system that contains a user-simulator dialogue bot to train the newly hired customer service agent, and an automated dialogue evaluation model to check the training result. The system combines different NLP technologies in a novel way to provide value for customer service.  

  \bibliographystyle{ACM-Reference-Format}
  \bibliography{main}

  \end{document}